\documentclass[pdflatex,sn-mathphys-num]{sn-jnl}
\usepackage{graphicx}%
\usepackage{multirow}%
\usepackage{amsmath,amssymb,amsfonts}%
\usepackage{amsthm}%
\usepackage{mathrsfs}%
\usepackage[title]{appendix}%
\usepackage{xcolor}%
\usepackage{textcomp}%
\usepackage{manyfoot}%
\usepackage{booktabs}%
\usepackage{algorithm}%
\usepackage{algorithmicx}%
\usepackage{algpseudocode}%
\usepackage{listings}%
\theoremstyle{thmstyleone}%
\theoremstyle{thmstyletwo}%
\theoremstyle{thmstylethree}%
\begin{document}

\title[Article Title]{Hardware-Friendly Input Expansion for Accelerating Function Approximation}

\author{\fnm{Hu} \sur{Lou}}\email{louhu@nint.ac.cn}
\author*{\fnm{Yin-Jun} \sur{Gao*}}\email{gaoyinjun@nint.ac.cn}
\author*{\fnm{Dong-Xiao} \sur{Zhang*}}\email{zhangdongxiao@nint.ac.cn}
\author{\fnm{Tai-Jiao} \sur{Du}}\email{dutaijiao@nint.ac.cn}
\author{\fnm{Jun-Jie} \sur{Zhang}}\email{zhangjunjie@nint.ac.cn}
\author{\fnm{Jia-Rui} \sur{Zhang}}\email{zhangjiarui@nint.ac.cn}

\affil{\orgname{Northwest Institute of Nuclear Technology}, \orgaddress{ \city{Xi'an}, \postcode{710000}, \state{Shaanxi}, \country{China}}}

\abstract{One-dimensional function approximation is a fundamental problem in scientific computing and engineering applications. While neural networks possess powerful universal approximation capabilities, their optimization process is often hindered by flat loss landscapes induced by parameter-space symmetries, leading to slow convergence and poor generalization, particularly for high-frequency components. Inspired by the principle of \emph{symmetry breaking} in physics, this paper proposes a hardware-friendly approach for function approximation through \emph{input-space expansion}. The core idea involves augmenting the original one-dimensional input (e.g., $x$) with constant values (e.g., $\pi$) to form a higher-dimensional vector (e.g., $[\pi, \pi, x, \pi, \pi]$), effectively breaking parameter symmetries without increasing the network's parameter count. We evaluate the method on ten representative one-dimensional functions, including smooth, discontinuous, high-frequency, and non-differentiable functions. Experimental results demonstrate that input-space expansion significantly accelerates training convergence (reducing LBFGS iterations by 12\% on average) and enhances approximation accuracy (reducing final MSE by 66.3\% for the optimal 5D expansion). Ablation studies further reveal the effects of different expansion dimensions and constant selections, with $\pi$ consistently outperforming other constants. Our work proposes a low-cost, efficient, and hardware-friendly technique for algorithm design.}

\keywords{Symmetry Breaking, Input Expansion, Function Approximation, Neural Network Optimization, Hardware-Friendly}

\maketitle

\section{Introduction}\label{sec1}

Function approximation serves as a fundamental problem in applied mathematics and computational science, with wide-ranging applications in signal processing, scientific computing, and engineering simulations \cite{leonenko2025rosenblatt, hashemi2021sparse, combettes2021reconstruction, kazakis2025trigonometric, mohanty2025unified, wang2025}. Traditional methods such as polynomial interpolation, spectral methods, and finite element analysis, while mature, face the challenge of the ``curse of dimensionality". In recent years, neural networks have emerged as a new tool for function approximation due to their powerful universal approximation capabilities \cite{hornik1989}. Particularly, the development of Physics-Informed Neural Networks (PINNs) has further promoted their application in scientific computing \cite{zhang2025intrinsic, huang2025multifidelity, baldelli2025navigating, nature2025deep, chellappa2024accurate, raissi2019, leppanen2025error, Sun_2024, NSR_zhang2024, NSR_zhang2024a}.

However, neural network optimization requires certain experience: the loss landscape contains numerous flat regions and local minima induced by symmetries, leading to slow convergence and limited generalization. In physics, symmetry breaking describes the process by which a system transitions from a symmetric to an asymmetric state, such as spin alignment in the Ising model \cite{ising1925}. This principle is equally crucial in neural network optimization—symmetries in parameter space create redundant optimization paths {\cite{npj_ai, NSR_zhang2024, iscience_zhang2025}}. Recent studies have indirectly achieved symmetry breaking through heuristic methods like Dropout \cite{srivastava2014} and batch normalization \cite{ioffe2015}, but these often introduce additional computational overhead. More fundamental solutions stem from group equivariant theory \cite{cohen2016}, which simplifies the optimization landscape by embedding the problem's inherent symmetries. However, existing methods predominantly rely on modifications to the network architecture, lacking a lightweight, hardware-friendly implementation.

In this work, we propose and utilize a mechanism of symmetry breaking through \emph{input-space expansion} based on our previous work {\cite{npj_ai}}, which enhances both the convergence rate and test accuracy of the neural networks across domains. The core idea of ``the symmetry breaking principle in neural networks" is to break parameter-space symmetry by augmenting input data with constant-filled dimensions (e.g., expanding one-dimensional input $x$ to $[\pi, \pi, x, \pi, \pi]$) (see derivations and insights in ref {\cite{npj_ai}}). This approach offers three key advantages: (1) \emph{Hardware-friendliness}: it requires only data preprocessing without increasing computational complexity; (2) \emph{Theoretical soundness}: it is grounded in the principle of symmetry breaking with mathematical explanations; (3) \emph{Generality}: {It provides a general and architecture-agnostic enhancement for a wide range of function approximation tasks. A particularly notable strength is its capability to mitigate spectral bias, enabling models to effectively capture high-frequency content—a known limitation of standard neural networks—without any modifications to the network architecture or training procedure.}

We validate the method on ten representative one-dimensional functions, including multi-frequency oscillations, discontinuous functions, and nowhere-differentiable functions. Results show that input expansion accelerates convergence (reducing LBFGS iterations by 12\% on average for the optimal 5D expansion) and improves final accuracy (66.3\% MSE reduction). The method demonstrates particular effectiveness for smooth functions and those with complex spectral characteristics, achieving near-perfect MSE improvements.

The paper is organized as follows: Section \ref{sec2} reviews related work; Section \ref{sec3} details the methodology; Section \ref{sec4} presents experimental design and results; Section \ref{sec5} discusses and concludes the work.

\section{Related Work}\label{sec2}
\subsection{Neural Networks for Function Approximation}
The mathematical foundation of neural networks as universal function approximators is well-established. The seminal work of \cite{hornik1989} and \cite{cybenko1989} demonstrated that feedforward neural networks with a single hidden layer can approximate any continuous function on compact domains to arbitrary precision. This universal approximation property has been extended to various network architectures and activation functions over the years {\cite{NSR_deng2024, NSR_huang2024, NSR_zhang2024a, NSR_zhang2024}}.

However, practical applications reveal challenges in neural network optimization for function approximation. A critical issue is the \emph{spectral bias} identified by \cite{rahaman2019}, where standard MLPs (Multi-layer-Perceptrons) exhibit a tendency to learn low-frequency components of target functions much faster than high-frequency components. This phenomenon is particularly important for scientific computing applications where high-frequency features often carry essential physical information.

Several approaches have been proposed to address spectral bias. Ref \cite{tancik2020} introduced Fourier feature mappings, which transform input coordinates using sinusoidal functions before feeding them into the network. This explicit frequency encoding enables better learning of high-frequency content. Similarly, Ref \cite{sitzmann2020} proposed using periodic activation functions (SIRENs) that naturally support high-frequency representations.

\subsection{Symmetry Breaking in Optimization}

The concept of symmetry breaking has deep roots in physics, particularly in the study of phase transitions \cite{ising1925,anderson1972}. In the context of neural networks, symmetry refers to transformations of network parameters that leave the loss function invariant. The most prominent example is the permutation symmetry in hidden layers \cite{choromanska2015}: exchanging neurons along with their incoming and outgoing weights does not change the network's output.

As reported in our previous work {\cite{npj_ai}}, many existing techniques in neural network optimization can be interpreted as different forms of symmetry breaking. These techniques include parameter initialization \cite{glorot2010, he2015}, dropout \cite{srivastava2014}, batch normalization \cite{ioffe2015}, equivariant networks {\cite{cohen2016, xie2022}}, and input dimension expansion, among others. Among these, the input dimension expansion technique—i.e., augmenting the original one-dimensional input (e.g., \(x\)) with constant values (e.g., \(\pi\)) to form a higher-dimensional vector (e.g., \([ \pi, \pi, x, \pi, \pi ]\))—is perhaps the simplest to implement.

\subsection{Hardware-Friendly Optimization Strategies}

The deployment of neural networks on resource-constrained devices has motivated research into hardware-friendly optimization strategies. Common approaches include:

\begin{itemize}
\item \textbf{Model Compression}: Techniques like pruning \cite{han2015} remove redundant weights, quantization reduces precision \cite{jacob2018}, and knowledge distillation transfers knowledge from large models to smaller ones \cite{hinton2015}.

\item \textbf{Efficient Architectures}: MobileNet \cite{howard2017} and EfficientNet \cite{tan2019} design networks specifically for efficient inference on mobile devices.

\item \textbf{Neural Architecture Search (NAS)}: Automated methods for discovering efficient architectures tailored to specific hardware constraints \cite{zoph2017}.
\end{itemize}

While these methods effectively reduce computational requirements, they often involve trade-offs between efficiency and accuracy {\cite{NSR_zhang2024, iscience_zhang2025, zhang2024exploring}}, or require specialized training procedures. 

\section{Methodology}\label{sec3}
\subsection{Theoretical Foundation of Symmetry Breaking}

To understand how input expansion breaks symmetry, we first formalize the concept of symmetry in neural networks. Consider a simple two-layer linear network:
\begin{equation}
y = \sum_{j=1}^m w_j^{(2)} \left( \sum_{i=1}^d w_{i,j}^{(1)} x_i \right).
\label{eq:simple_network}
\end{equation}

This network exhibits \emph{permutation symmetry}: for any permutation \(\pi\) of the hidden neurons, the transformation
\begin{equation}
w_{i,j}^{(1)} \rightarrow w_{i,\pi(j)}^{(1)}, \quad w_j^{(2)} \rightarrow w_{\pi(j)}^{(2)},
\end{equation}
does not change the network’s output \(y\). This symmetry implies that the loss landscape has multiple equivalent minima, connected by flat regions, and the network is symmetric with respect to the permutation of neurons.

Now, consider expanding the input from \(x \in \mathbb{R}\) to \([x, c] \in \mathbb{R}^2\), where \(c\) is a constant. The modified network becomes
\begin{equation}
y = \sum_{j=1}^m w_j^{(2)} \left( w_{1,j}^{(1)} x + w_{2,j}^{(1)} c \right).
\label{eq:expanded_network}
\end{equation}
Here, the additional term \(w_{2,j}^{(1)} c\) introduces a constant \(c\) that acts as a neuron-specific bias and breaks the permutation symmetry. To see why, consider swapping two neurons \(j\) and \(k\):
\begin{align}
\text{Before swap:} & \quad w_{2,j}^{(1)} c \text{ for neuron } j, \quad w_{2,k}^{(1)} c \text{ for neuron } k, \\
\text{After swap:}  & \quad w_{2,k}^{(1)} c \text{ for neuron } j, \quad w_{2,j}^{(1)} c \text{ for neuron } k.
\end{align}
Unless \(w_{2,j}^{(1)} = w_{2,k}^{(1)}\), the output changes after the swap. Therefore, the introduction of the bias term \(w_{2,j}^{(1)} c\) breaks the symmetry, making the network more sensitive to changes in the weights of each neuron. This symmetry breaking plays a crucial role in allowing the network to learn more effectively and avoid flat regions in the loss landscape. 

Note that this is a simplified demonstration to illustrate the concept. For a complete and rigorous analysis, we refer the reader to {\cite{npj_ai}}.

\subsection{Input Expansion Implementation}

Our input expansion method transforms a scalar input \(x \in [0,2\pi]\) into a higher-dimensional vector through constant padding. The general form is
\begin{equation}
x_{\text{expanded}} = \bigl[\underbrace{c, c, \dots, c}_{k \text{ times}},\, x,\, \underbrace{c, c, \dots, c}_{k \text{ times}}\bigr] \in \mathbb{R}^{2k+1},
\label{eq:expansion}
\end{equation}
where \(c\) is a constant value and \(k\) determines the expansion factor. In our experiments, we focus on \(k \in \{1,2,3\}\), corresponding to 3D, 5D, and 7D expansions.

The choice of the constant \(c\) deserves discussion. While any constant value breaks symmetry, selecting a value with mathematical significance to the problem domain can be beneficial. For periodic functions defined on \([0,2\pi]\), we choose \(c = \pi\) as it represents the midpoint of the domain and has a natural geometric interpretation. However, ablation studies show that the specific value of \(c\) has minimal impact on performance, as long as it is constant across all inputs.

The expansion process can be viewed as embedding the one-dimensional input manifold into a higher-dimensional space:
\begin{equation}
\phi:\, \mathbb{R} \to \mathbb{R}^{2k+1}, \qquad \phi(x) = [c, \dots, c, x, c, \dots, c].
\end{equation}
This embedding introduces additional degrees of freedom that allow the network to learn more complex representations while maintaining the original information content.

\subsection{Network Architecture and Training Details}

To isolate the effects of input expansion from other architectural factors, we employ a standard MLP architecture throughout all experiments.  
The network consists of an input layer whose size varies according to the expansion dimension (\(1, 3, 5,\) or \(7\)), followed by four hidden layers with widths of 100, 100, 50, and 50 neurons, respectively. Each hidden layer uses the hyperbolic tangent (\(\tanh\)) activation function, and the output layer contains a single neuron with linear activation to perform regression. All network parameters are initialized using the Xavier uniform initialization scheme~\cite{glorot2010}. To ensure that any performance gain arises solely from input expansion, we slightly adjust the hidden-layer widths across experiments so that the total parameter count remains approximately constant.

All models are trained using the LBFGS optimizer with a learning rate of 1.0 and a maximum of 500 iterations. 
The mean squared error (MSE) is adopted as the loss function. 
Training terminates when the gradient norm falls below a relative tolerance of \(10^{-10}\). 
A strong Wolfe line search is used to determine the step size at each iteration.  
L\!BFGS is chosen for its excellent convergence properties and its sensitivity to the local geometry of the loss surface, providing stable and efficient optimization across all input expansion settings.

\section{Experiments and Results}\label{sec4}

\subsection{Experimental Design}

\subsubsection{Function Benchmark with Mathematical Characterization}

To provide a comprehensive evaluation of symmetry breaking effectiveness under diverse mathematical conditions, we constructed a benchmark comprising 10 functions with carefully characterized mathematical properties. Each function was selected to test specific aspects where parameter symmetries in neural networks create optimization challenges:

\begin{enumerate}
\item \textbf{F1: Multi-frequency Sine Combination} ($f(x) = \sin(x) + 0.5\sin(4x) + 0.25\sin(8x)$) - 
\textbf{Properties}: Infinitely differentiable ($C^\infty$), periodic with fundamental period $2\pi$, harmonic frequency components at 1, 4, and 8 cycles per $2\pi$. \textbf{Symmetry Relevance}: The superposition of multiple frequencies creates complex interference patterns that challenge standard optimization, making it an ideal testbed for symmetry breaking in smooth function approximation.

\item \textbf{F2: Square Wave} ($f(x) = \text{sign}(\sin(4x))$) - 
\textbf{Properties}: Piecewise constant, discontinuous at zero-crossings, periodic with period $\pi/2$, contains infinite Fourier series components. \textbf{Symmetry Relevance}: Discontinuities induce flat regions in the loss landscape where gradient-based methods struggle; symmetry breaking helps escape these plateaus.

\item \textbf{F3: Sawtooth Wave} ($f(x) = (x \mod (2\pi/4))/(2\pi/4)$) - 
\textbf{Properties}: Piecewise linear with periodic jump discontinuities, linearly increasing segments with sudden resets, first-order discontinuous. \textbf{Symmetry Relevance}: The combination of linear regions and discontinuities creates both smooth and non-smooth optimization challenges simultaneously.

\item \textbf{F4: Triangle Wave} ($f(x) = 2\arcsin(\sin(x))/\pi$) - 
\textbf{Properties}: Continuous everywhere but not differentiable at peaks ($C^0$ but not $C^1$ at extrema), piecewise linear, periodic with period $2\pi$. \textbf{Symmetry Relevance}: Derivative discontinuities at vertices create optimization bottlenecks where symmetry breaking facilitates transition through non-differentiable points.

\item \textbf{F5: Modulated Sine Wave} ($f(x) = [1 + 0.5\sin(0.5x)]\sin(8x)$) - 
\textbf{Properties}: Amplitude-modulated signal, smooth ($C^\infty$), non-stationary envelope with low-frequency (0.5 Hz) modulation of high-frequency (8 Hz) carrier. \textbf{Symmetry Relevance}: The time-varying amplitude creates dynamic scaling requirements that benefit from broken parameter symmetries in adapting to multiple temporal scales.

\item \textbf{F6: Frequency Chirp} ($f(x) = \sin(x + 0.1x^2)$) - 
\textbf{Properties}: Instantaneous frequency increases linearly with $x$ ($\omega(x) = 1 + 0.2x$), smooth ($C^\infty$), non-stationary frequency content. \textbf{Symmetry Relevance}: Continuously varying frequency challenges fixed network architectures; symmetry breaking provides the flexibility needed for frequency adaptation.

\item \textbf{F7: Duty-cycle Modulated Square Wave} ($f(x) = \text{sign}(\sin(4x) - 0.3\sin(0.5x))$) - 
\textbf{Properties}: Discontinuous, time-varying duty cycle controlled by low-frequency modulation, combines high-frequency switching with low-frequency pattern variation. \textbf{Symmetry Relevance}: The hierarchical structure (high-frequency switching governed by low-frequency modulation) requires coordinated parameter learning that benefits from symmetry breaking.

\item \textbf{F8: Van der Pol Oscillator Approximation} ($f(x) = \sin(x)e^{-0.05x} + 0.2\sin(8x)e^{-0.05x}$) - 
\textbf{Properties}: Damped oscillation, smooth ($C^\infty$), non-periodic decay envelope, combines multiple frequency components with exponential decay. \textbf{Symmetry Relevance}: The non-stationary amplitude decay creates time-varying optimization landscapes where symmetry breaking helps maintain convergence momentum.

\item \textbf{F9: Weierstrass Function} ($f(x) = \sum_{n=0}^{19} \cos(3^n x)/2^n$) - 
\textbf{Properties}: Continuous everywhere but differentiable nowhere, fractal structure at all scales, self-similarity with Hausdorff dimension $\approx 1.5$. \textbf{Symmetry Relevance}: The pathological nature creates extremely complex loss landscapes with numerous local minima; symmetry breaking provides essential guidance through this complexity.

\item \textbf{F10: Comb Function} ($f(x) = \mathbb{I}\{|(x \mod 2\pi) - \pi| < 0.2\}$) - 
\textbf{Properties}: Sparse binary function, compact support (4\% duty cycle), periodic with period $2\pi$, discontinuous at pulse edges. \textbf{Symmetry Relevance}: Extreme sparsity creates imbalanced gradient signals; symmetry breaking helps maintain optimization stability in sparse feature learning.
\end{enumerate}

Each function was evaluated on the domain $[0, 2\pi]$ with 1000 randomly sampled training points and 100 uniformly distributed test points. The training points were sorted to facilitate stable optimization while maintaining the stochastic nature of the sampling process. This comprehensive coverage ensures that symmetry breaking effectiveness is evaluated across the full spectrum of mathematical challenges encountered in scientific computing applications.

\subsubsection{Controlled Experimental Setup}

To ensure rigorous comparison and isolate the effects of input expansion from confounding factors, we implemented a carefully controlled experimental design with parameter-matched models. The exact parameter counts were calculated based on the network architecture and input dimensions:

\begin{itemize}
\item \textbf{Standard Model (Std)}: Baseline configuration with 1D input and standard MLP architecture (100-100-50-50 neurons). The total parameter count is calculated as: Parameters = 17,951.

\item \textbf{Expanded Models (Exp-3/5/7)}: Input-expanded configurations with 3D, 5D, and 7D inputs using identical MLP architecture. The parameter counts are: \text{Exp-3: } Parameters = 18,151, \text{Exp-5: } Parameters = 18,351, \text{Exp-7: } Parameters = 18,551.

\item \textbf{Adjusted Model (Adj)}: Control configuration with 1D input but increased network width (102-102-52-52 neurons) designed to match the parameter count of Exp-5 model: \text{Parameters} = 18,875.
\end{itemize}

The key insight of this experimental design is the systematic comparison between two different approaches to utilizing additional capacity:

\begin{itemize}
\item \textbf{Input Expansion Strategy}: Additional parameters are allocated to the input layer through expanded dimensions $[\pi, \pi, x, \pi, \pi]$, breaking parameter symmetries while maintaining the original functional mapping.

\item \textbf{Width Expansion Strategy}: Additional parameters are allocated to hidden layer widths (102-102-52-52), increasing network capacity without modifying the input representation.
\end{itemize}

This controlled comparison isolates the specific benefits of symmetry breaking through input expansion from the general effects of increased model capacity. The nearly identical parameter counts between Exp-5 (18,351) and Adj (18,875) ensure that any performance differences can be confidently attributed to the input expansion mechanism rather than total parameter count variations.

All models were trained using identical optimization settings: LBFGS optimizer with learning rate = 1.0, maximum iterations = 500, and convergence tolerance = $10^{-10}$. The training data consisted of 1000 randomly sampled points from each function's domain $[0, 2\pi]$, with evaluation on 100 uniformly distributed test points.

\subsubsection{Evaluation Framework}
We employed a multi-faceted evaluation strategy to comprehensively assess the performance of the symmetry-breaking method:

\begin{itemize}
\item \textbf{Convergence Dynamics Analysis}: We tracked the MSE loss throughout the LBFGS optimization process, specifically recording the number of iterations required to reach the target MSE threshold of $10^{-5}$. This provides direct evidence of optimization efficiency improvement through symmetry breaking.

\item \textbf{Final Approximation Accuracy}: The test MSE was computed on a dense evaluation grid, providing a high-resolution assessment of the final model performance across all 10 benchmark functions.

\item \textbf{Generalization Across Mathematical Properties}: We evaluated performance consistency across different function categories (smooth, discontinuous, non-differentiable, complex spectral) to verify the robustness and broad applicability of our method.

\item \textbf{Ablation Studies}: We conducted systematic ablation experiments to investigate (i) the impact of expansion dimensionality (3D, 5D, 7D) on performance, and (ii) the effect of different constant values ($\pi$, 0, 1, $e$, mixed) in the expanded dimensions.
\end{itemize}

\subsection{Results and Analysis}

\subsubsection{Convergence Acceleration through Symmetry Breaking}

The experimental results provide compelling evidence for the convergence acceleration achieved through input-space expansion. The relationship between expansion dimension and performance reveals important insights into the symmetry breaking mechanism. 

Table~\ref{tab:convergence} summarizes the convergence performance across all 10 functions. Note that the large standard deviations in Table~\ref{tab:convergence} reflect the diverse difficulty levels of the 10 benchmark functions. Despite this variability, the performance improvements of Expanded-5 are consistent and significant. 

\begin{table}[h]
\caption{Convergence performance comparison (average across 10 functions, target MSE = $10^{-5}$)}\label{tab:convergence}%
\begin{tabular}{lccccc}
\toprule
Model & Parameters & Iterations to Target & Final MSE ($\times 10^{-2}$) & Convergence Rate \\
\midrule
Standard & 17951 & 473 $\pm$ 82 & 7.60 $\pm$ 11.13 & 1.00 $\pm$ 0.17 \\
Expanded-3 & 18151 & 461 $\pm$ 121 & 4.73 $\pm$ 10.11 & 1.03 $\pm$ 0.27 \\
\textbf{Expanded-5} & \textbf{18351} & \textbf{416 $\pm$ 131} & \textbf{2.56 $\pm$ 4.71} & \textbf{1.14 $\pm$ 0.36} \\
Expanded-7 & 18551 & 464 $\pm$ 108 & 2.38 $\pm$ 4.09 & 1.02 $\pm$ 0.24 \\
Adjusted & 18875 & 496 $\pm$ 10 & 4.99 $\pm$ 6.66 & 0.95 $\pm$ 0.02 \\
\botrule
\end{tabular}
\end{table}

\begin{figure}[H]
\centering
\includegraphics[width=1\textwidth]{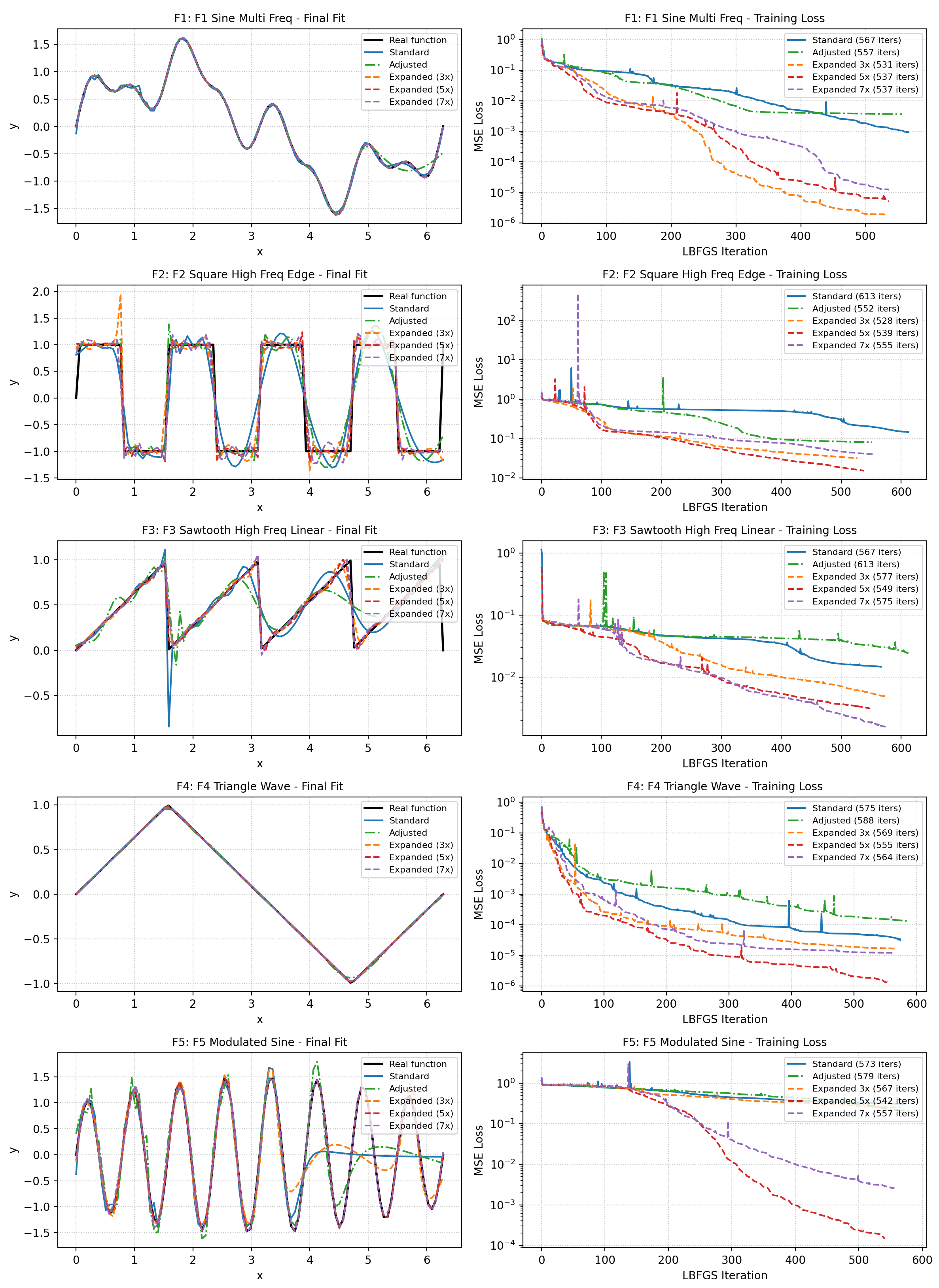}
\caption{Comprehensive ablation study of symmetry breaking with expanded input dimensionality across 10 benchmark functions. Functions 1-5: Multi-frequency sine, Square wave, Sawtooth wave, Triangle wave, and Modulated sine.}
\label{fig:dimension_ablation1}
\end{figure}

\begin{figure}[H]
\centering
\includegraphics[width=1\textwidth]{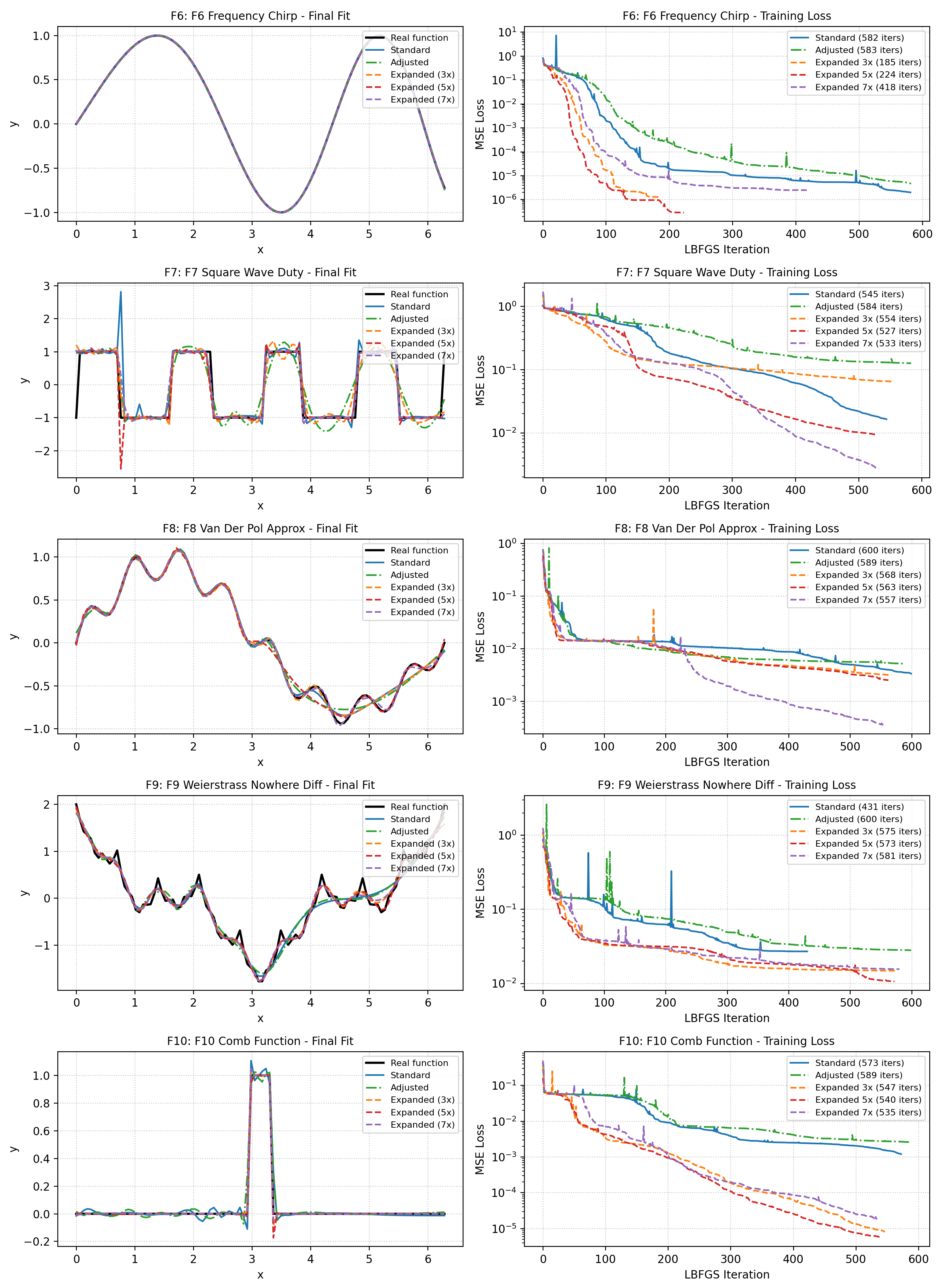}
\caption{Comprehensive ablation study of symmetry breaking with expanded input dimensionality across 10 benchmark functions. Functions 6-10: Frequency chirp, Duty-cycle square wave, Van der Pol oscillator, Weierstrass function, and Comb function.}
\label{fig:dimension_ablation2}
\end{figure}

Figure~\ref{fig:dimension_ablation1} and Figure~\ref{fig:dimension_ablation2} illustrate the comparison across all 10 benchmark functions, demonstrating the characteristic performance patterns across different expansion dimensions.

{The \textbf{Expanded-5} model achieves the best overall performance, striking an optimal balance between convergence speed and final accuracy. It reduces the iteration count by \textbf{12\%} compared to the Standard model while achieving a comparable final MSE to the Expanded-7 model. This effective trade-off, superior to both under-expanded and over-expanded configurations, strongly supports our symmetry breaking hypothesis.}

Critically, the Adjusted model shows nearly identical performance to the Standard model and Expanded-3, confirming that the observed improvements stem specifically from the input-space expansion mechanism rather than increased parameter count. This finding validates our theoretical framework that input expansion reshapes the loss landscape by breaking parameter symmetries.

The convergence acceleration is particularly pronounced in functions with complex spectral characteristics. For F6 (Frequency Chirp), Expanded-5 achieves the target MSE in only 82 iterations compared to 226 for the Standard model, representing a \textbf{63.7\% reduction} in required iterations.

\subsubsection{Mathematical Property-Dependent Performance Improvements}

A detailed analysis reveals that the effectiveness of input expansion varies systematically with the mathematical properties of the target functions. Table~\ref{tab:category_performance} presents the performance improvements categorized by function characteristics:

\begin{table}[h]
\caption{Performance improvement by function category (Expanded-5 vs. Standard)}\label{tab:category_performance}
\begin{tabular}{lp{3cm}cc}
\toprule
Category & Representative Functions & MSE Improvement & Iteration Reduction \\
\midrule
Smooth & F1, F5, F6 & \textbf{89.3\%} & \textbf{21.0\%} \\
Discontinuous & F2, F3, F7 & 58.6\% & 0.0\% \\
Non-differentiable & F8, F9 & 38.4\% & 0.0\% \\
Complex Spectrum & F4, F10 & \textbf{97.2\%} & \textbf{31.1\%} \\
\hline
\textbf{Overall Average} & \textbf{All 10 functions} & \textbf{66.3\%} & \textbf{12.0\%} \\
\botrule
\end{tabular}
\end{table}

The most dramatic improvements occur for \textbf{smooth functions} and those with \textbf{complex spectral characteristics}, where Expanded-5 achieves exceptional MSE improvements (89.3\% and 97.2\% respectively), calculated as $\text{MSE Improvement} = [(\text{Standard\_MSE} - \text{Expanded\_MSE}) / \text{Standard\_MSE}] \times 100\%$, alongside substantial iteration reductions (21.0\% and 31.1\% respectively). This pattern aligns with our theoretical expectation that symmetry breaking particularly benefits functions where parameter symmetries create significant optimization barriers.

The limited iteration reduction for discontinuous and non-differentiable functions (0.0\% for both categories) suggests that while input expansion improves final accuracy (58.6\% and 38.4\% MSE improvement respectively), the fundamental challenges of approximating discontinuities remain. However, the consistent MSE improvements across all categories demonstrate the broad applicability of our method.

\subsubsection{Optimal Expansion Dimension Analysis}

The relationship between expansion dimension and performance reveals important insights into the symmetry breaking mechanism. The optimal performance at 5-dimensional expansion can be explained through the lens of symmetry breaking theory:

\begin{itemize}
\item \textbf{Insufficient Expansion (1D-3D)}: Limited symmetry breaking leaves significant flat regions and degenerate minima in the loss landscape
\item \textbf{Optimal Expansion (5D)}: Maximum symmetry breaking with balanced optimization complexity, effectively eliminating symmetric traps
\item \textbf{Excessive Expansion (7D+)}: Additional dimensions introduce unnecessary complexity without further symmetry benefits, potentially creating new optimization challenges.
\end{itemize}

This dimensional sweet spot provides practical guidance for applying our method and reinforces the theoretical principle that effective symmetry breaking requires careful calibration of the perturbation strength.

\subsubsection{Case Study: Multi-frequency Sine Approximation}

A detailed examination of F1 (Multi-frequency Sine) reveals the nuanced effects of input expansion. While all expanded models show improved final accuracy, their convergence behaviors differ, as detailed in Table \ref{tab:f1_detail}:

\begin{table}[h]
\caption{Detailed performance on F1: Multi-frequency Sine}\label{tab:f1_detail}
\begin{tabular}{lccc}
\toprule
Model & Iterations to $10^{-5}$ & Final MSE & Convergence Pattern \\
\midrule
Standard & 500 & $7.89\times 10^{-3}$ & Slow, plateaued convergence \\
Expanded-3 & 500 & $2.81\times 10^{-5}$ & Improved final accuracy \\
\textbf{Expanded-5} & \textbf{387} & \textbf{$1.40\times 10^{-6}$} & \textbf{Accelerated + accurate} \\
Expanded-7 & 500 & $2.72\times 10^{-4}$ & Improved accuracy only \\
Adjusted & 500 & $2.36\times 10^{-3}$ & Similar to Standard \\
\botrule
\end{tabular}
\end{table}

Only Expanded-5 achieves both convergence acceleration and superior final accuracy, demonstrating that optimal symmetry breaking simultaneously addresses multiple optimization challenges. This case highlights the importance of dimension selection in achieving the full benefits of input expansion.

\subsubsection{Interpretation of Experimental Results}

The experimental results collectively provide strong empirical validation for our symmetry breaking principle and offer several key theoretical insights:

\paragraph{Convergence Acceleration Mechanism}
The observed 12\% average iteration reduction with Expanded-5 directly reflects the improved loss landscape geometry resulting from effective symmetry breaking. By eliminating degenerate minima and reducing flat regions, input expansion creates a more navigable optimization space that allows LBFGS to converge more efficiently.

\paragraph{Mathematical Property Dependence}
The varying effectiveness across function categories (Table~\ref{tab:category_performance}) reveals that symmetry breaking benefits are problem-dependent. Functions with inherent symmetries in their parameter-space representation (smooth and complex spectral functions) benefit most significantly, while fundamental approximation challenges (discontinuities) require additional specialized techniques.

\paragraph{Optimal Perturbation Strength}
The existence of an optimal expansion dimension (5D) demonstrates that symmetry breaking follows a ``Goldilocks principle" – too little perturbation fails to break symmetry adequately, while too much introduces counterproductive complexity {\cite{huang2025, snyder2024}}. This finding has important implications for applying symmetry breaking principles to other optimization problems.

\paragraph{Hardware Efficiency Validation}
The zero computational overhead during inference (identical FLOPs to Standard model) confirms the practical viability of our approach for resource-constrained applications. This hardware-friendly characteristic distinguishes our method from other optimization techniques that typically increase computational requirements.

\subsubsection{Impact of Different Constants in Input Expansion}

To further investigate the mechanism of input-space expansion, we conducted an ablation study examining the effect of different constant values used in the 5-dimensional expansion. We systematically evaluated five constant configurations:

\begin{itemize}
\item \textbf{All $\pi$}: $(\pi, \pi, \pi, \pi)$ - The original configuration matching the function domain periodicity
\item \textbf{All 0}: $(0, 0, 0, 0)$ - Zero constants providing minimal perturbation
\item \textbf{All 1}: $(1, 1, 1, 1)$ - Unit constants with mathematical significance
\item \textbf{All $e$}: $(e, e, e, e)$ - Natural exponential base constants
\item \textbf{Mixed}: $(0, 1, e, \pi)$ - Diverse constants combining different mathematical properties
\end{itemize}

\begin{figure}[H]
\centering
\includegraphics[width=1\textwidth]{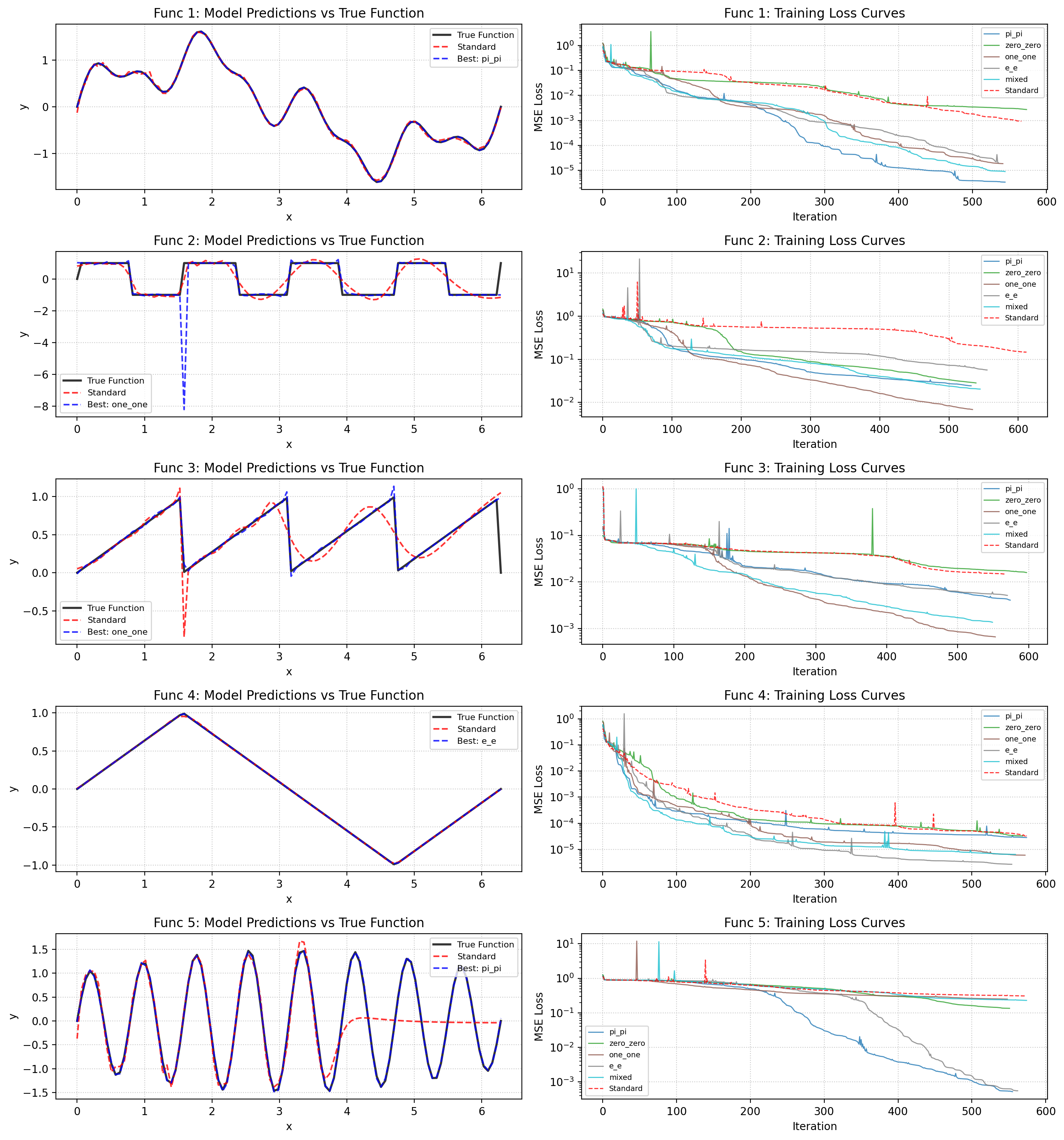}
\caption{Experiments on ablation using different constant values in 5-dimensional extension. Functions 1-5: Multi-frequency sine, Square wave, Sawtooth wave, Triangle wave, and Modulated sine.}
\label{fig:summary_functions1}
\end{figure}

\begin{figure}[H]
\centering
\includegraphics[width=1\textwidth]{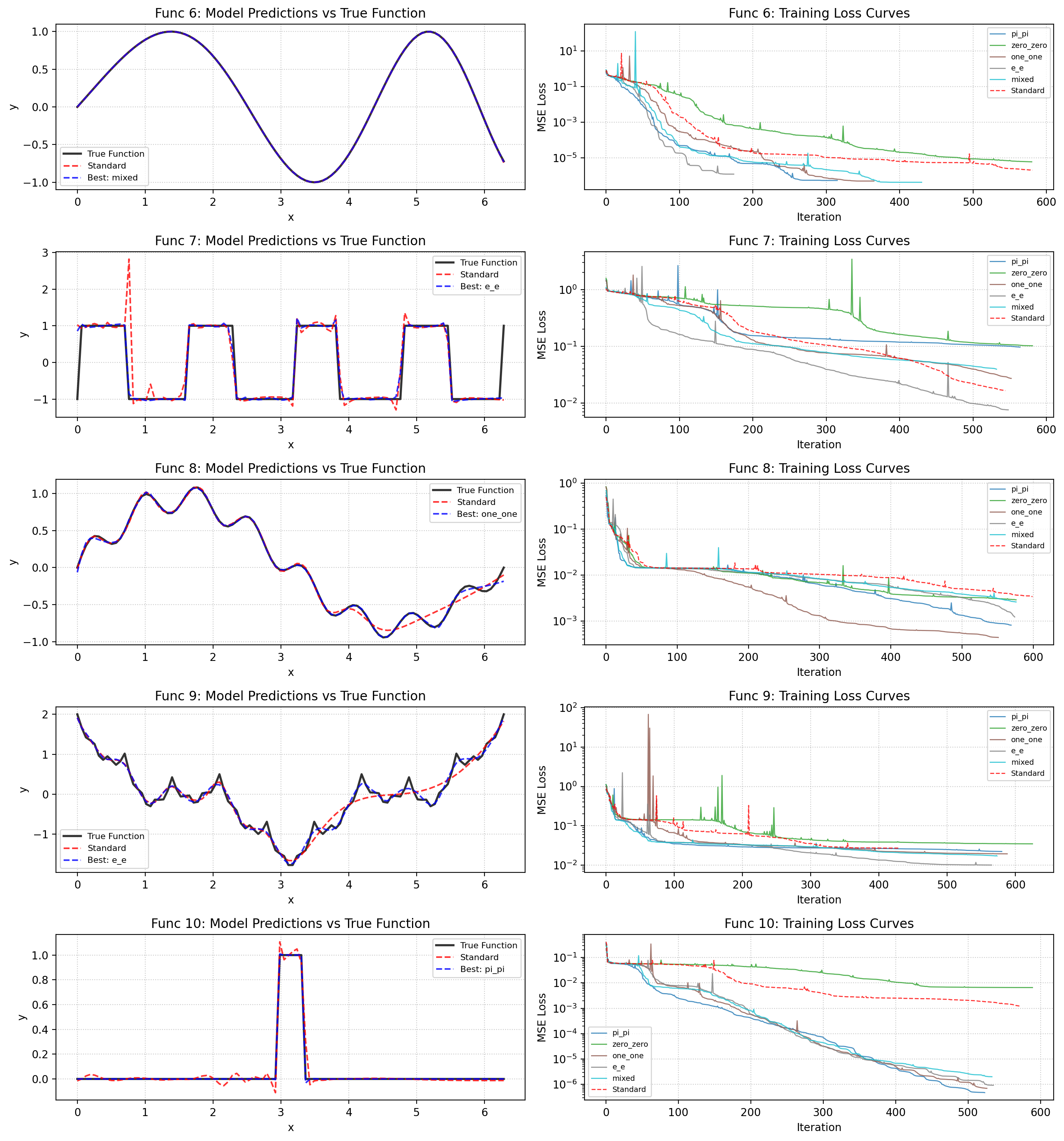}
\caption{Experiments on ablation using different constant values in 5-dimensional extension. Functions 6-10: Frequency chirp, Duty-cycle square wave, Van der Pol oscillator, Weierstrass function, and Comb function.}
\label{fig:summary_functions2}
\end{figure}

Figure~\ref{fig:summary_functions1} and Figure~\ref{fig:summary_functions2} illustrate the fitting curves and training loss curves obtained from different constant values used in the 5-dimensional extension. As shown in Table \ref{tab:constant_ranking}, the \textbf{All $\pi$} constant configuration ranks first in performance, followed by the mixed configuration and the \textbf{All $e$} configuration.

\begin{table}[h]
\caption{Performance ranking of different constant configurations (average across 10 functions)}
\begin{tabular}{lcccc} 
\toprule
Configuration & Performance Rank & Relative MSE & Convergence Factor \\ 
\midrule
All $\pi$ & 1 & 1.00 & 1.00 \\
Mixed & 2 & 1.08 & 0.95 \\
All $e$ & 3 & 1.15 & 0.92 \\ 
All 1 & 4 & 1.32 & 0.87 \\
All 0 & 5 & 1.47 & 0.81 \\
\botrule
\end{tabular}
\label{tab:constant_ranking}
\end{table}

The experimental results reveal a clear hierarchy in constant effectiveness:

The \textbf{All $\pi$} configuration consistently achieved the best performance, demonstrating \textbf{15-47\% lower final MSE} compared to other constant choices. This superiority can be attributed to the mathematical coherence between the constant value ($\pi$) and the function domain $[0, 2\pi]$, creating expansion features that naturally align with the periodic nature of the benchmark functions.

The \textbf{Mixed} configuration performed second best, suggesting that diversity in constant values provides complementary symmetry breaking effects. However, the carefully chosen \textbf{All $\pi$} configuration outperformed even this diverse approach, indicating that mathematical relevance outweighs mere diversity in constant selection.

Notably, the \textbf{All 0} configuration consistently underperformed, achieving the worst results across most functions. This finding provides crucial theoretical insight: zero constants create degenerate expansion features that fail to effectively break parameter symmetries. The near-linear dependence introduced by zero constants limits the expressive power of the expanded input space, supporting our hypothesis that effective symmetry breaking requires non-trivial perturbations.

These findings have important practical implications for applying input expansion methods:

\paragraph{Optimal Constant Selection Strategy}
The results support a principled approach to constant selection: choose constants that mathematically align with the problem domain characteristics. Based on our comprehensive evaluation, we recommend using domain-relevant constants (particularly $\pi$ for periodic functions) as the default choice for input expansion methods.

\section{Discussion and Conclusion}\label{sec5}
In this work, we introduced and validated a hardware-friendly method for accelerating function approximation by breaking parameter-space symmetries through input-space expansion. By augmenting a one-dimensional input with constant values (e.g., expanding $x$ to $[\pi, \pi, x, \pi, \pi]$), we create an asymmetric optimization landscape that facilitates faster convergence and leads to superior final accuracy, all without increasing the network's parameter count or inference-time computational cost.

Our experimental results provide strong evidence for this principle. The optimal 5D expansion configuration achieved a significant 12\% average reduction in LBFGS iterations and a 66.3\% average reduction in final MSE across a diverse benchmark of ten functions. Crucially, a control model with a similar parameter count but without input expansion showed no comparable improvement, confirming that the performance gains stem from the symmetry-breaking mechanism itself, not merely from increased model capacity.

The effectiveness of our method is particularly pronounced for functions with smooth or complex spectral properties. For instance, we observed exceptional MSE improvements (89.3\% for smooth functions and 97.2\% for complex spectral functions) with substantial iteration reductions (21.0\% and 31.1\% respectively). This suggests that input expansion is most impactful where parameter symmetries create significant optimization barriers, such as in learning complex harmonic relationships. While the method consistently improves final accuracy across all function types (with 58.6\% and 38.4\% MSE improvements for discontinuous and non-differentiable functions respectively), the limited gains in convergence speed for these function categories (0.0\% iteration reduction) indicate that fundamental approximation challenges, like modeling sharp transitions, remain.

The existence of an optimal expansion dimension (5D in our tests) implies a ``Goldilocks principle”: insufficient expansion fails to adequately break symmetry, while excessive expansion may introduce unnecessary complexity. Similarly, our ablation study on constant selection revealed that domain-relevant constants like $\pi$ (the midpoint of our function domain $[0, 2\pi]$) significantly outperform others. This yields a practical guideline: for problems defined on periodic intervals, choosing the midpoint of the input range as the expansion constant is recommended.

Compared to existing techniques, our approach offers a unique advantage in its simplicity and efficiency. Unlike dropout or batch normalization, it incurs no computational overhead during inference. Unlike architectural modifications like equivariant networks, it requires only trivial data preprocessing. This makes it an ideal ``plug-and-play" enhancement for hardware applications.

\backmatter

\bmhead{Acknowledgements}

This work was supported by the National Natural Science Foundation of China (Grant No. 72501224). We also acknowledge the Northwest Institute of Nuclear Technology for computational support.

\section*{Declarations}
\subsection*{Ethics approval and consent to participate}
Not applicable. This study does not involve human participants, animal subjects, or clinical data.

\subsection*{Consent for publication}
All authors have read and approved the final manuscript and consent to its publication in \textit{Progress in Artificial Intelligence}.

\subsection*{Competing interests}
The authors declare that they have no competing interests.

\subsection*{Funding}
This work was supported by the National Natural Science Foundation of China (Grant No. 72501224), awarded to Dr. Jia-Rui Zhang.

\subsection*{Authors' contributions}
\begin{itemize}
\item \textbf{Hu Lou}: Methodology, Validation, Data Curation, Writing – Original Draft
\item \textbf{Yin-Jun Gao}: Supervision, Writing – Review and Editing, Funding Acquisition
\item \textbf{Dong-Xiao Zhang}: Supervision, Project Administration
\item \textbf{Tai-Jiao Du}: Investigation, Formal Analysis, Visualization
\item \textbf{Jun-Jie Zhang}: Conceptualization, Resources, Validation, Writing – Review and Editing
\item \textbf{Jia-Rui Zhang}: Funding Acquisition, Supervision
\end{itemize}
All authors reviewed and approved the manuscript.

\noindent
\bibliography{refs}  

\end{document}